\documentclass{article}

\usepackage[final]{corl_2021} 

\usepackage{times}
\usepackage{multicol}
\usepackage{wrapfig}
\usepackage{graphicx}
\usepackage{amsmath} 
\usepackage{amssymb} 
\usepackage{pifont}
\usepackage{soul,color} 
\usepackage{mathtools}
\usepackage{booktabs}
\usepackage{multirow}
\usepackage{enumerate}
\usepackage{enumitem}
\usepackage{balance}
\usepackage{gensymb}
\usepackage{tabulary}
\usepackage{dblfloatfix}
\usepackage{algorithm,setspace}
\usepackage{algpseudocode}
\usepackage{lipsum}
\usepackage{siunitx}
\usepackage{amsthm}
\usepackage{units}
\usepackage{colortbl}
\usepackage{cancel}

\usepackage{afterpage}
\newlength{\oldtextfloatsep}

\usepackage[font=footnotesize,labelfont=bf]{caption}

\graphicspath{{figures/}}

\newcommand{\figGap}[0]{\vspace{-1.1\baselineskip}}
\definecolor{lightgreen}{rgb}{0.8,1,0.8}

\makeatletter
\def\th@plain{%
  \thm@notefont{}
  \itshape 
}
\def\th@definition{%
  \thm@notefont{}
  \normalfont 
}

\makeatletter
\def\blfootnote{\gdef\@thefnmark{}\@footnotetext}
\makeatother

\makeatother

\newtheoremstyle{indented}{3pt}{3pt}{\addtolength{\leftskip}{2.5em}}{}{\bfseries}{.}{.5em}{}
\theoremstyle{indented}
\theoremstyle{definition}

\newcommand{\argminx}{\underset{{x}}{\operatorname{argmin}}}

\title{LEO: Learning Energy-based Models in \\ Factor Graph Optimization}

\author{
  Paloma Sodhi\textsuperscript{1,2}, Eric Dexheimer\textsuperscript{1}, Mustafa Mukadam\textsuperscript{2}, Stuart Anderson\textsuperscript{2}, Michael Kaess\textsuperscript{1} \\
  \textsuperscript{1}Carnegie Mellon University, \textsuperscript{2} Facebook AI Research\\
}

\begin{document}
\maketitle


\vspace{-6mm}
\begin{abstract}

We address the problem of learning observation models end-to-end for estimation. Robots operating in partially observable environments must infer latent states from multiple sensory inputs using observation models that capture the joint distribution between latent states and observations. This inference problem can be formulated as an objective over a graph that optimizes for the most likely sequence of states using all previous measurements. Prior work uses observation models that are either known a-priori or trained on surrogate losses independent of the graph optimizer. In this paper, we propose a method to directly optimize end-to-end tracking performance by learning observation models with the graph optimizer in the loop. This direct approach may appear, however, to require the inference algorithm to be fully differentiable, which many state-of-the-art graph optimizers are not. Our key insight is to instead formulate the problem as that of energy-based learning. We propose a novel approach, LEO, for learning observation models end-to-end with graph optimizers that may be non-differentiable. LEO alternates between sampling trajectories from the graph posterior and updating the model to match these samples to ground truth trajectories. We propose a way to generate such samples efficiently using incremental Gauss-Newton solvers. We compare LEO against baselines on datasets drawn from two distinct tasks: navigation and real-world planar pushing. We show that LEO is able to learn complex observation models with lower errors and fewer samples.

\end{abstract}


\vspace{-3mm}
\section{Introduction}
\label{sec:introduction}

\blfootnote{Code and supplementary material can be found on \url{https://psodhi.github.io/leo}}

We focus on the problem of learning observation models end-to-end for estimation. Consider a robot hand manipulating an object with only tactile feedback. It must reason over \emph{a sequence of} touch observations over time to collapse uncertainty about the latent object pose. A common way to solve this is as an inference over a factor graph which relies on observation models that can map between states and observations \cite{dellaert2020factor, dellaert2017factor,cadena2016past}. However, in many domains, sensors that produce observations are complex and difficult to model. \emph{Can we instead learn observation models from data?}


Given a batch of ground truth trajectories and corresponding measurements, how should we learn observations models? One approach would be to learn a direct mapping from measurement to state, for example as a regression or classification problem \cite{sodhi2021tactile, baikovitz2021gpr,sundaralingam2019robust,lambert2018deep}. However, while easy to optimize given a direct supervised loss, this only minimizes a surrogate loss independent of the graph optimizer and is not guaranteed to minimize the final tracking errors that we care about. The other option would be to directly minimize final tracking errors between optimized and ground truth trajectories \cite{yi2021differentiable, jatavallabhula2020slam}. However, many state-of-the-art factor graph optimizers, e.g.\ iSAM2 \cite{kaess2012isam2}, are not natively differentiable due to operations such as dynamic re-linearizations\footnotemark. In such cases, we are limited to black-box search for learning parameters which is very sample inefficient \cite{hansen2019pycma, hu2015covariance, snoek2012practical}.

\begin{figure}[!t]
	\centering
	\includegraphics[width=\columnwidth, trim=0 0.5mm 0 0, clip=true]{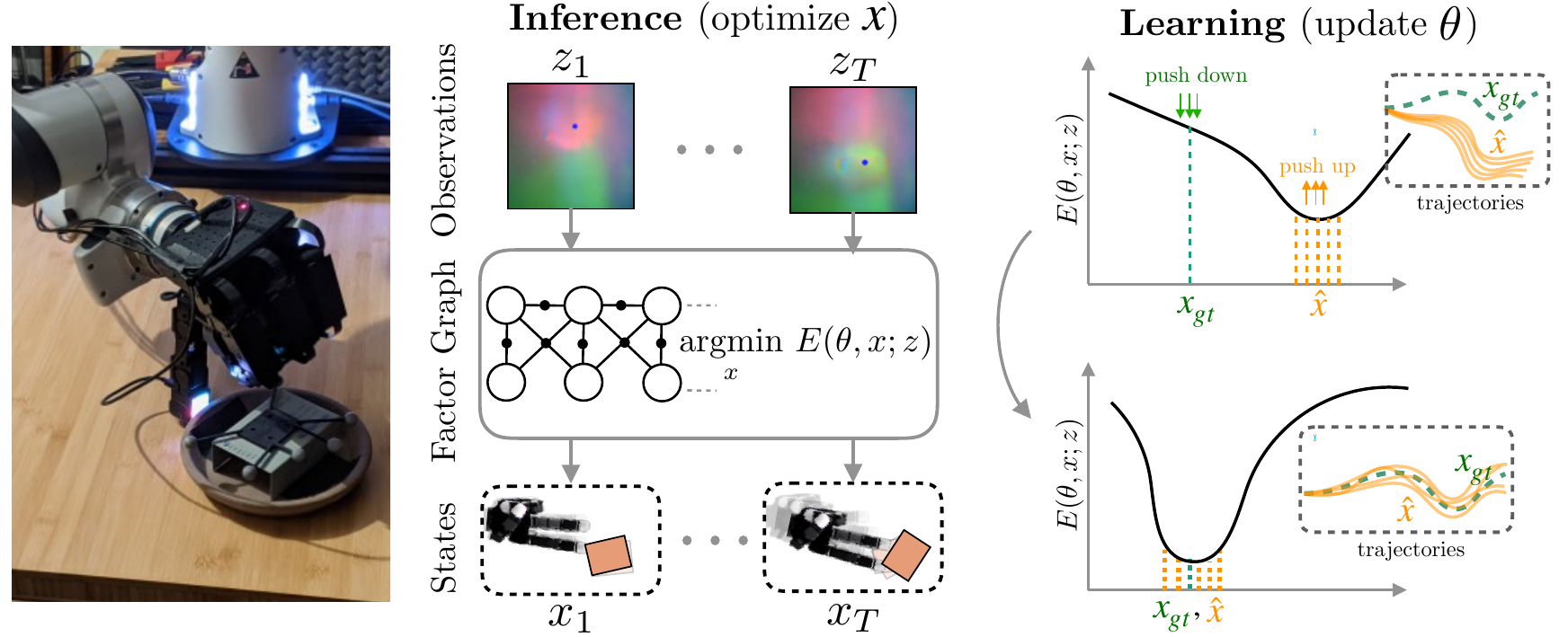}
	\caption{We show that learning observation models can be viewed as shaping energy functions that graph optimizers, even non-differentiable ones, optimize. \textbf{Inference} solves for most likely states $x$ given model and input measurements $z$. \textbf{Learning} updates observation model parameters \(\theta\) from training data.}
	\label{fig:coverInferenceLearning}
	\figGap
\end{figure}

Instead of differentiating through the optimization process, we note that what we ultimately care about is the final solution from the optimizer, which depends \emph{only on the shape} of the optimized cost function. We would like a cost function that has low cost around the observed ground truth trajectories and high cost elsewhere. This is precisely what energy-based models aim to do by shaping an ``energy'' function to be low around observed data and high elsewhere~\cite{lecun2006tutorial, kumar2019maximum, belanger2017end}. We cast our problem of learning observation models for estimation as energy-based learning.

We propose a novel approach, \textbf{L}earning \textbf{E}nergy-based Models in Factor Graph \textbf{O}ptimization (LEO), for learning observation models end-to-end with graph optimizers that may be non-differentiable. LEO alternates between sampling trajectories from the graph posterior and updating the model to match the distribution of samples to ground truth trajectories. We show that we can generate such samples efficiently using incremental Gauss-Newton graph solvers that create a Gaussian approximation for the Boltzmann distribution defined over a continuous space of trajectories.
\footnotetext{See Appendix A for discussion on applying LEO to differentiable optimizers.}
To the best of our knowledge, this is the first paper to show that even non-differentiable factor graph models can be learned efficiently. Our main contributions are:
\begin{enumerate}[topsep=0pt, leftmargin=0.75cm]
	\item A novel formulation of learning observation models in estimation as energy-based learning.
	\item An algorithm, LEO, to learn observation models that minimize end-to-end tracking errors in factor graph optimizers that may be non-differentiable.
	\item Empirical evaluation on two distinct tasks: synthetic navigation and real-world planar pushing.
\end{enumerate}

\vspace{-3mm}
\section{Related Work}
\label{sec:relatedwork}
\vspace{-1mm}

\textbf{Filtering and smoothing based estimation.}
Observation models represent joint or conditional distributions between states and measurements. Early estimation methods utilized these models within filtering contexts such as Kalman filters and EKFs/UKFs~\cite{julier2004unscented, thrun2001robust}. Recent methods have also looked at making these filters differentiable for end-to-end learning~\cite{kloss2020train, lee2020multimodal, jonschkowski2018differentiable, venkatraman2017predictive, sun2016learning, haarnoja2016backprop}. However, filtering can be inconsistent for nonlinear estimation problems due to linearization based on past, marginalized states that cannot be undone~\cite{julier2001counter}. Instead, problems in simultaneous localization and mapping (SLAM) are increasingly formulated as smoothing or nonlinear optimization objectives~\cite{cadena2016past, dellaert2017factor, dellaert2020factor}. Unlike filtering, smoothing allows dynamic relinearization of past states as well as exploits the inherent sparsity resulting in efficient and more accurate solutions. Factor graphs are an increasingly popular way for solving such smoothing problems~\cite{dellaert2020factor, dellaert2017factor, sodhi2021tactile, baikovitz2021gpr, yi2021differentiable, ortiz2021visual, czarnowski2020deepfactors, hartley2018hybrid, fourie2016nonparametric}, offering a flexible modeling framework while being efficient to optimize. We use the factor graph based smoothing framework in this paper.


\textbf{Learning for smoothing.}
Typically, observation models used in factor graphs are analytic models defined a-priori~\cite{mur2017orb, engel2014lsd, lambert2019joint}. Recent work has looked at using learned models~\cite{sodhi2021tactile, baikovitz2021gpr, sundaralingam2019robust} or learned measurement representations~\cite{czarnowski2020deepfactors, bloesch2018codeslam} either to be used independently or to be plugged into a graph optimizer. However, these model parameters are learned on surrogate losses independent of the graph optimizer, and hence do not directly attempt to minimize final tracking errors.

An alternative approach is to view observation model parameters as hyper-parameters of the graph and perform a hyper-parameter search. This is usually done as sampling and derivative-free approaches using black-box solvers~\cite{hansen2019pycma, hu2015covariance, snoek2012practical}. However, the high computational burden associated with sampling and evaluation can be prohibitive for large parameter spaces.


\textbf{Differentiable optimization.}
Instead of a search, one may learn these model parameters via a continuous gradient-based optimizer matching graph optimizer output against a desired solution. Minimizing this loss would, however, require differentiating through the argmin optimization process. One way to do this is via \emph{unrolling} where the optimization is represented as a series of differentiable, gradient based updates~\cite{belanger2017end, wilder2019end, bechtle2021meta}. Since smoothing is a nonlinear least-squares optimization, it too can be unrolled in a similar fashion via differentiable Levenberg-Marquardt updates~\cite{jatavallabhula2020slam, lv2019taking, tang2018ba} or differentiable Gauss-Newton updates~\cite{yi2021differentiable, von2020gn, bhardwaj2020dgpmp2}. However, vanilla unrolling suffers from some drawbacks, notably the learned cost function can be sensitive to the specific optimization procedure, e.g. number of unrolling steps \cite{amos2020differentiable}. Moreover, this approach cannot be readily applied to commonly used optimizers relying on non-differentiable heuristics such as those in GTSAM, SNOPT \cite{dellaert2012factor, gill2005snopt}. An alternate to unrolling is to employ the implicit function theorem which states that an optimal energy function must satisfy first-order necessary conditions \cite{dontchev2009implicit, amos2017optnet}. While this is promising for convex problems, for non-convex problems, one may recover an energy function with multiple minima such that the optimizer at test time may find itself in a different basin from ground truth.



\textbf{Energy-based learning.}
Instead of differentiating through the optimization process, we note what we ultimately care about is the final solution from the optimizer, which depends only on the shape of the cost function. This is what energy-based models aim to do by shaping an ``energy'' function to be low around observed data and high elsewhere~\cite{lecun2006tutorial}. Energy-based modeling is a family of unsupervised learning methods popular in many structured prediction tasks such as inverse reinforcement learning~\cite{ziebart2008maximum, du2019ebm, wulfmeier2015maximum, kitani2012activity} and computer vision~\cite{gustafsson2020train, du2020improved}. Recently, a number of approaches have looked at training deep energy-based models~\cite{chen2015learning, belanger2016structured, belanger2017end}. The challenge with such methods is generating samples from the energy based model. Methods either rely on MCMC sampling~\cite{tieleman2008training, du2019implicit, nijkamp2019learning} which does not scale well with dimension or resort to learning a separate generator~\cite{kim2016deep, dai2017calibrating, kumar2019maximum}. We propose a way to generate such samples efficiently using incremental Gauss-Newton solvers. Recent work~\cite{yoon2021unsupervised, wong2020variational} has also looked at learning observation models using variational inference techniques when ground truth states are not available. However, in our work, we can assume some ground truth demonstrations from a motion capture system which makes the problem more tractable.

\vspace{-3mm}
\section{Problem Formulation}
\label{sec:probform}
\vspace{-2mm}

We formalize our problem of learning observation models under the general theoretical framework of energy based models. We begin by describing the observation model, how it is optimized at inference time and finally how its parameters are learned from training data (Fig. \ref{fig:coverInferenceLearning}).

\vspace{-3mm}
\subsection{Model}
\label{sec:probform:model}

We define an energy-based model that captures dependencies between variables in a graph by associating a scalar energy $E(\cdot)$ to each configuration of the graph. Our goal is to model the likelihood of a sequence of latent states $x$ given a sequence of measurements $z$. We adopt a factor graph framework (Fig. \ref{fig:coverInferenceLearning}(a)) for expressing this likelihood as\footnote{We use the word likelihood as in typical learning formulations. Specifically, it refers to the posterior of states given measurements. Model parameters $\theta$ include covariances $\Sigma_k$ in the norm $||\cdot||_{\Sigma_k}$ as well.},
\begin{equation}
    \small
	\begin{split}
	\label{eq:probform:eq3.1.1}
	P_{\theta}(x|z) & = \frac{1}{\mathcal{Z}(\theta;z)}\exp\left\{-E(\theta, x; z)\right\} = \frac{1}{\mathcal{Z}(\theta;z)}\exp\left\{-\frac{1}{2}\sum_{k}||f({\theta,x_k;z_k})||^2\right\}
\end{split}
\end{equation}
where, $x:=\{x_1 \dots x_{T}\}$ is the sequence of latent states, and $z:=\{z_1 \dots z_{T}\}$ is the sequence of measurements. $||f({\theta,x_k;z_k})||^2$ are the factor costs with $f({\theta,x_k};z_k)$ being the local observation model mapping a \emph{subset} of states $x_k\subseteq x$ and measurements $z_k\subseteq z$ to a cost, $\theta$ are the learnable model parameters, and $\mathcal{Z}(\theta; z)=\int_{x}\exp (-E(\theta, x; z))dx$ is the normalization or partition function.

Note that Eq. \ref{eq:probform:eq3.1.1} is a special case of the Boltzmann distribution where the energy $E(\cdot)$ is a sum-of-squares. Notably, when $f({\theta,x_k};z_k)$ is linearized, $E(\cdot)$ is quadratic in $x$ and subsequently $P_{\theta}(x|z)$ becomes a Gaussian distribution. This is the structure that Gauss-Newton solvers employ, and we will leverage this fact to efficiently sample trajectories as described later in Section \ref{sec:approach:gaussnewton}.

\vspace{-3mm}
\subsection{Inference}
\label{sec:probform:inference}

At inference time, given a sequence of measurements $z$, we wish to solve for the most likely sequence of latent states $x$. We formulate this as maximizing the log-likelihood of the model,
\begin{equation}
	\begin{split}
	\label{eq:probform:eq3.1.2}
	\hat{x} & = \underset{{x}}{\operatorname{argmax}}\ \log P_{\theta}(x|z) \\
	\end{split}
\end{equation}

Substituting the model expression from Eq. \ref{eq:probform:eq3.1.1}, and dropping constants $\mathcal{Z}(\theta;z)$ and $\nicefrac{1}{2}$, results in a nonlinear least-squares minimization of the form,
\begin{equation}
\small
\begin{split}
\label{eq:probform:eq3.1.3}
\hat{x} & = \underset{{x}}{\operatorname{argmax}}\ \log \frac{1}{\mathcal{Z}(\theta;z)}\exp\left\{-\frac{1}{2}\sum_{k}||f({\theta,x_k};z_k)||^2\right\} = \underset{{x}}{\operatorname{argmin}}\sum_{k}^{}||f({\theta,x_k};z_k)||^2 \\
\end{split}
\end{equation}
We use an efficient online Gauss-Newton graph solver for this objective. This enables both real-time inference as well as efficient sampling for learning model parameters during training (Section \ref{sec:approach:gaussnewton}).

\vspace{-3mm}
\subsection{Learning}
\label{sec:probform:learning}


At train time, our goal is to learn a model that explains training data of pairs $(x_{gt},z)$. We express this as minimizing a loss $\mathcal{L}(\theta)$ over the training data
\begin{equation}
	\begin{split}
	\label{eq:probform:eq3.3.1}
	\mathcal{L}(\theta) =  \frac{1}{|\mathcal{D}|}\sum_{(x_{gt}^{i},z^{i})\in \mathcal{D}} \mathcal{L}(\theta; x^{i}_{gt}, z^{i}) \\
	\end{split}
\end{equation}
where, $\{x^{i}_{gt}, z^{i}\}\in \mathcal{D}$ is a training dataset of ground truth trajectories and measurements. 

There are various choices of losses $\mathcal{L}(\cdot)$ for learning model parameters $\theta$,

\textbf{Loss 1: Minimize final tracking loss.}
A straightforward choice for $\mathcal{L}(\cdot)$ is to directly minimize final tracking errors between optimized $\hat{x}^{i}$ and ground truth ${x^{i}_{gt}}$ trajectories, i.e.\ $\mathcal{L}(\theta; x_{gt}, z)=\nicefrac{1}{|\mathcal{D}|}\sum_{i}||\hat{x}^{i}\ominus x^{i}_{gt}||^2_2$. While this loss directly minimizes final tracking errors, it may not be differentiable since the graph inference path from $\theta$ to $\hat{x}$ is not differentiable. In such cases, we are limited to black-box hyperparameter search which is very sample inefficient.

\textbf{Loss 2: Minimize surrogate loss.}
An alternate choice for the loss $\mathcal{L}(\cdot)$ is to directly map measurements to ground truth states, for example as a regression or classification problem. Such a surrogate loss is independent of graph inference, and unlike Loss 1, is differentiable and easy to optimize. However, this does not minimize the final tracking errors that we care about.

\textbf{Loss 3: Minimize energy-based loss.}
Finally, we consider a loss that is a function of the energy, i.e.\ $\mathcal{L}(E(\theta, \cdot); x_{gt}^{i}, z^{i})$. Intuitively, this assigns a low loss to \textit{well-behaved} energy functions, i.e.\ functions that give the lowest energy to training data of ground truth trajectories (correct answers) and higher energy to unseen data (incorrect answers).

This energy-based loss, unlike Loss 2, is highly correlated with the final tracking loss as it shapes the energy (or cost) landscape so as to make the inference step return trajectories closer to the ground truth. Moreover, unlike Loss 1, this loss is only a function of energies which is differentiable \emph{irrespective of inference}. Hence, we use Loss 3 and detail our approach in the next section.

\vspace{-5pt}
\section{Approach}
\label{sec:approach}


We instantiate the broad framework of energy-based learning within the context of factor graph estimation in 4.1 and 4.2. In 4.3 we show how incremental Gauss Newton approaches enable efficient sampling from these energy based models.

\vspace{-3mm}
\subsection{Normalized Negative Log-Likelihood Loss}
\label{sec:approach:nll}

We define the overall loss as the negative log-likelihood (NLL) of the data under the energy model\footnote{The motivation for NLL loss comes from probabilistic modeling, notably maximum entropy moment matching. We provide a detailed derivation along with its Gaussian approximation in Appendix C.}:
\begin{equation}
	\begin{split}
	\label{eq:approach:eq4.1.4}
	\mathcal{L}(\theta) & =  \frac{1}{|\mathcal{D}|}\sum_{(x_{gt}^{i},z^{i})\in \mathcal{D}} -\log P_{\theta}(x_{gt}^{i}|z^{i}) \\
\end{split}
\end{equation}


Substituting $P_\theta(x|z)$ as the Boltzman distribution expression from Eq. \ref{eq:probform:eq3.1.1} we have,
\begin{equation}
	\begin{split}
	\label{eq:approach:eq4.1.5}
	\mathcal{L}(\theta) =  \frac{1}{|\mathcal{D}|}\sum_{(x_{gt}^{i},z^{i})\in \mathcal{D}} E(\theta; x_{gt}^{i}, z^{i}) + \log \int_x \exp(-E(\theta; x, z^{i}))dx \\
	\end{split}
\end{equation}

The expression above has an intuitive explanation. The first term penalizes high energy over ground truth training samples. The second term penalizes low energy everywhere else, thus acting as a ``contrastive'' term. This has a desirable effect of globally shaping the energy function so that the minima lies at the training samples. The contrastive term prevents the energy surface from becoming flat and thus degenerate \cite{lecun2006tutorial}. Unfortunately, we cannot directly plug in Eq. \ref{eq:approach:eq4.1.5} as a standard machine learning loss (e.g.\ in PyTorch) to optimize. This is because the second log partition term is over a continuous space of trajectories $x$, which renders exact computation intractable for arbitrary nonlinear cost functions \cite{wainwright2005new}.

Instead, let us take the gradient of this loss, and substitute $P_{\theta}(x|z^{i})$ expression from Eq. \ref{eq:probform:eq3.1.1},
\begin{equation}
	\begin{split}
	\label{eq:approach:eq4.1.8}
	\nabla_{\theta} \mathcal{L}(\theta) = \frac{1}{|\mathcal{D}|}\sum_{(x_{gt}^{i},z^{i})\in \mathcal{D}} \left[ \underbrace{\nabla_{\theta} E(\theta; x_{gt}^{i}, z^{i}))}_\text{ground truth samples} - \underset{\substack{x\sim \\ P_{\theta}(x|z)}}{\operatorname{\mathbb{E}}}\underbrace{\nabla_{\theta} E(\theta; x, z^{i}))}_\text{learned distribution samples}\right]
	\end{split}
\end{equation}


We see the log partition term transforms into an expectation over the learned distribution. The gradient expression above has a similar interpretation as the loss. The first term pushes down the energy of ground truth training samples. Interestingly, the second term now pulls up the energy of samples drawn from the learned Boltzmann distribution. If one is able to generate such samples, computing the gradient is straightforward. 

Generally, sampling from the continuous Boltzmann distribution is intractable, relying typically on MCMC methods. But as we will detail in Section~\ref{sec:approach:gaussnewton}, we can indeed generate such samples efficiently since our graph optimizer maintains a Gaussian approximation of this distribution. Additionally in practice, we found that introducing a temperature term $T$ inside the Boltzmann distribution, i.e.\  $P_\theta(x|z) \propto \exp(-\nicefrac{1}{T}E({\theta},x;z))$, was critical for controlling the variance of the samples. Setting $T=0$ corresponds to only using the mode of the distribution which in this case is the mean trajectory. This corresponds to minimizing a generalized perceptron loss~\cite{lecun2006tutorial}.

\begin{figure}[!t]
	\centering
	\includegraphics[width=\textwidth, trim = 0cm 0cm 0cm 0cm]{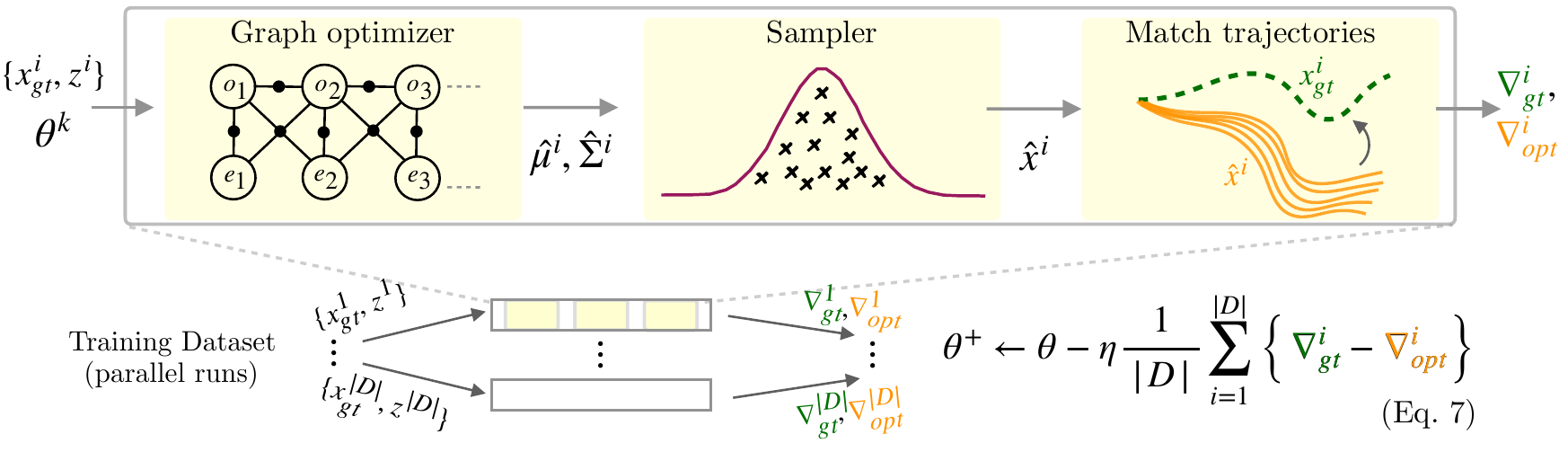}
	\caption{LEO algorithm: For each training example, the zoomed in panel shows the three main stages: (a) solve graph optimization, (b) sample trajectories from graph posterior, and (c) compute gradients. Gradients are then pooled together to update parameters $\theta$.}
	\label{fig:approachOverall}
	\figGap
\end{figure}

\setlength{\textfloatsep}{0pt} 
	\begin{algorithm}[!t]
	    \scriptsize
		\caption{LEO algorithm}
		\label{algo:leo}
		\begin{algorithmic}
			\State \textbf{Input} Initial $\theta$, Training data $\{x^{i}_{gt}$, $z^i\}^{|\mathcal{D}|}_{i=1}$
			\While{\texttt{<!converged>}}
			\For{each training example i}
			\State $\hat{\mu}^{i}, \hat{\Sigma}^{i}=\underset{{x}}{\operatorname{argmin}}\{E(\theta, x; z^{i})\}$ \Comment{Solve graph optimizer objective}
			\State $\hat{x}^{i} \sim \mathcal{N}(\hat{\mu}^{i}, \hat{\Sigma}^{i})$ \Comment{Sample $S$ trajectories from graph posterior}
			\EndFor
			\State $\theta^{+}\leftarrow\theta^{}-\eta\frac{1}{|\mathcal{D}|}\sum\limits_{i=1}^{|\mathcal{D}|}\left\{\nabla_{\theta}E(\theta; x^{i}_{gt}, z^{i})-\frac{1}{{S}}\sum\limits_{\hat{x}^{i}}^{}\nabla_{\theta}E(\theta; \hat{x}^{i}, z^{i})\right\}$
			\Comment{Update model parameters (Eq. \ref{eq:approach:eq4.1.8})}
			\EndWhile
		\end{algorithmic}
  \afterpage{\global\setlength{\textfloatsep}{\oldtextfloatsep}}
	\end{algorithm}

\vspace{-3mm}
\subsection{LEO Algorithm}
\setlength{\textfloatsep}{18pt} 

Algorithm \ref{algo:leo} summarizes the LEO algorithm, also illustrated in Fig.~\ref{fig:approachOverall}. It takes as inputs the training data containing ground truth state and measurement pairs, i.e. $\{x^{i}_{gt}$, $z^i\}^{|\mathcal{D}|}_{i=1}$, along with initial values for the learnable observation model parameters $\theta$. The goal is to converge to a $\theta$ that  minimizes the log-likelihood loss in Eq. \ref{eq:approach:eq4.1.4} over the given training set $i=1\dots |\mathcal{D}|$.

LEO invokes a black-box, non-differentiable graph optimizer iSAM2 in the loop to generate samples. Within each LEO iteration, for every training example $i$, we invoke the graph optimizer to obtain a Gaussian approximation $\mathcal{N}(\hat{\mu}^{i}, \hat{\Sigma}^{i})$ to the original Boltzmann distribution $P_{\theta}(x|z)$ about its mode. We then sample trajectories $\hat{x}^{i} \sim \mathcal{N}(\hat{\mu}^{i}, \hat{\Sigma}^{i})$ from this distribution.

The next step is to compute gradients of the energy, i.e.\ $\nabla_\theta E(\theta; {x}_{gt}^{i}, z^{i})$ and $\nabla_\theta E(\theta; \hat{x}^{i}, z^{i})$, at the ground truth $x^{i}_{gt}$ and the samples $\hat{x}^{i}$, respectively. Note that the gradients here are only w.r.t.\ $\theta$, and not $x$, and hence do not require the graph optimizer to be differentiable. Once we have the two gradients, we use the difference between them to update the observation model parameters $\theta$. These updated parameters are then used in the next LEO iteration.

\vspace{-3mm}
\subsection{Efficient Inference via Incremental Gauss-Newton}
\label{sec:approach:gaussnewton}






The only thing that remains to be specified in Algorithm \ref{algo:leo} is --- How do we generate samples efficiently from the graph inference? In Section \ref{sec:probform:inference} we noted that our inference algorithm must satisfy two requirements (a) real-time inference and (b) sampling from the Boltzmann distribution. We achieve this using via incremental Gauss-Newton to solve the nonlinear least squares in Eq. \ref{eq:probform:eq3.1.3}.

Gauss-Newton proceeds by linearizing observation model functions $f(\cdot)$ about a point $x_k^{0}$,
\begin{equation}
	\begin{split}
		\label{eq:probform:eq3.1.4}
		f({\theta,x_k};z_k)=f(\theta, x_k^0+\delta x_k; z_k)\approx f(\theta, x_k^0; z_k)+F_k\delta {x_k}
	\end{split}
\end{equation}
where, ${x_k}^0$ is the linearization point, $F_k=\left.\frac{\partial f({\theta, x_k;z_k})}{\partial {x_k}}\right|_{x_k^0}$ is the measurement Jacobian, and $\delta {x_k}=x_k-x_k^0$ the state update vector. Measurements $z$ and learnable parameters $\theta$ are treated as constant within the graph optimization step. Substituting the Taylor expansion in Eq. \ref{eq:probform:eq3.1.4} back into Eq. \ref{eq:probform:eq3.1.3},
\begin{equation}
\begin{split}
\label{eq:probform:eq3.1.5}
\delta x^* &= \underset{\delta{x}}{\operatorname{argmin}}\sum_{k}^{}||F_k\delta{x}_k-\left(-f(\theta, x_k^0; z_k)\right)||^2 = \underset{\delta{x}}{\operatorname{argmin}}\ ||A\delta{x}-b||^2_2
\end{split}
\end{equation}
where, $A$, $b$ are constructed by concatenating all $F_k$, $f(\theta, x_k^{0}; z_k)$ together  into a single matrix and vector respectively. We then iterate $x^{0}\leftarrow x^{0}+\delta x^*$ until convergence.


We can solve for the objective in Eq.~\ref{eq:probform:eq3.1.5} efficiently by exploiting two properties of the problem: \emph{sparsity} and \emph{online} measurements. For SLAM problems, both $A$ and $A^TA$ are large but sparse matrices, and can hence be factorized efficiently using sparse QR or Cholesky factorization respectively \cite{dellaert2006square}. However, this still does not guarantee real-time inference during test time since $A$ grows with measurements over time, making the method increasingly slower. Instead, subsequent incremental solvers \cite{kaess2012isam2, kaess2008isam, sodhi2020ics} reuse matrix factorizations from previous time steps. In this work, we use iSAM2 \cite{kaess2012isam2}, an incremental Gauss-Newton graph solver whose resulting MAP solution can be expressed as a Gaussian probability distribution with mean and covariance,
\begin{equation}
	\begin{split}
	\label{eq:probform:eq3.1.7}
	\hat{\mu} = \argminx \sum_{k}^{}||f({\theta, x_k;z_k})||^2, \hspace{30pt} 
	\hat{\Sigma} = \left(A^TA\right)^{-1}|_{x=\hat{\mu}}
	\end{split}
\end{equation}
This is the Gaussian approximation of the original Boltzmann distribution $P_{\theta}(x|z^{i})$ which we can now sample from efficiently. The graph posterior covariance $\hat{\Sigma}$ for such a Gauss-Newton solver is obtained using just Jacobians $A$ without needing to compute a Hessian \cite{dellaert2017factor}. Note that sampling $\hat{x}\sim \mathcal{N}(\hat{\mu}, \hat{\Sigma})$ is done at the tangent space of the linearized graph, i.e.\ $\hat{x}\leftarrow\hat{\mu}\oplus\delta x$ where ${\delta x}\sim \mathcal{N}(0, \hat{\Sigma})$ and $\oplus$ is a retraction \cite{sola2018micro}. The Gauss-Newton method for inference is an efficient technique for sampling from intractable models. A recent overview of the Laplace approximation with Gauss-Newton method can be found in \cite{immer2021improving}. Here, we build a specific Laplace approximation using the incremental solver iSAM2 that scales to sampling efficiently in real-time from very large graphs.\footnote{See Appendix B for run time comparison to an alternate MCMC sampler.}

\vspace{-3mm}
\section{Results and Evaluation}
\label{sec:results}

We evaluate LEO on synthetic navigation and real-world planar pushing. We compare against baselines on metrics like final tracking error and sample efficiency. We implement LEO in PyTorch \cite{paszke2019pytorch} and interface it with the GTSAM C++ library \cite{dellaert2012factor}, which contains several factor graph optimizers. We use the iSAM2 \cite{kaess2012isam2} optimizer for real-time, online optimization.

\begin{figure}[!t]
	\figGap
	\centering
	\includegraphics[width=0.9\columnwidth]{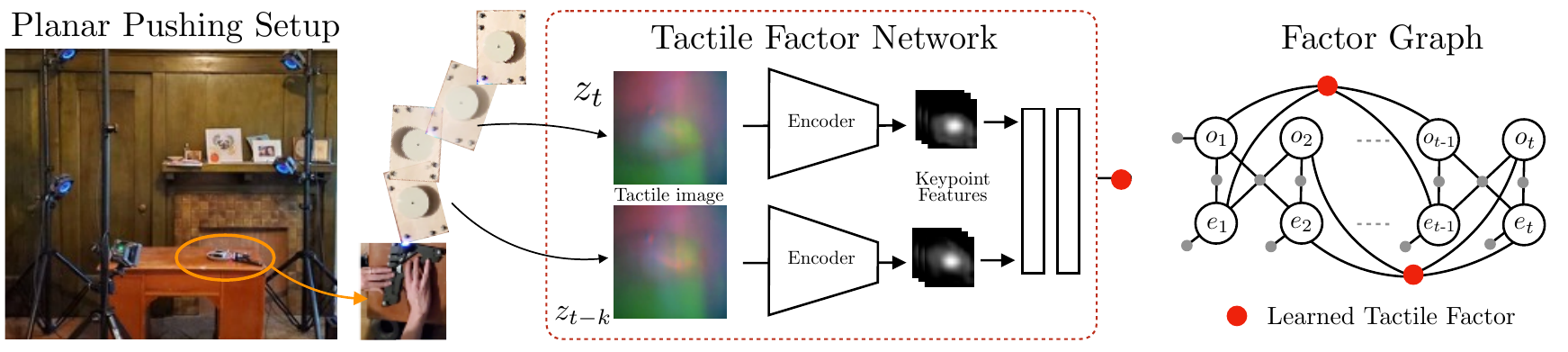}
	\caption{\textbf{Real-world planar pushing setup:} Learned tactile factors penalize deviations between predicted relative pose from the tactile factor network and estimated relative pose using variables in the graph. Tactile factor network contains an image to keypoint feature encoder pre-trained using a self-supervised loss (and kept fixed) \cite{sodhi2021tactile} and a keypoint to relative transform network that is fine-tuned on the end-to-end tracking errors.}
	\figGap
\label{fig:experimentalSetupPlanarPushing}
\end{figure}

\subsection{Overview}

\textit{Experimental setup:} For synthetic navigation (Datasets N1--N4), the goal is to estimate latent robot poses $x_t \in SE(2)$ from relative odometry and absolute GPS measurements. For real-world planar pushing (Datasets P1--P3), the goal is to estimate latent object poses $o_t \in SE(2)$ from tactile image measurements using a DIGIT tactile sensor \cite{lambeta2020digit} as shown in the setup in Fig. \ref{fig:experimentalSetupPlanarPushing}. Both these problems are modeled as factor graphs with poses as variable nodes and measurements as factor nodes.

\textit{Observation model parameters:} Recall factor costs are defined by the local observation model $f(\cdot)$ and learned model parameters $\theta$. Since we use Gaussian factors, we will learn their mean and covariance parameters, $\theta:=\{\mu, \Sigma\}$. For the navigation task, we learn (a) fixed covariances $\theta:=\{\Sigma_{odom},\Sigma_{gps}\}$ (N1, N2), and (b) covariances as a function of measurements $\theta:=\{\Sigma_{odom}(z), \Sigma_{gps}(z)\}$ (N3, N4). For the planar pushing task, we learn (a) fixed covariances $\theta:=\{\Sigma_{tac}, \Sigma_{qs}\}$ (P1, P2), and (b) means as a function of tactile image measurements $\theta:=\{\mu_{tac}(z)\}$ (P3). For all experiments, we use a diagonal covariance model.


\textit{Baselines:} We compare against a learned sequence model such as LSTM, and black-box hyperparameter search methods such as CMAES \cite{hansen2019pycma} and Nelder-Mead, all of which optimize Loss 1. We also compare against a surrogate supervised learning method \cite{sodhi2021tactile} which optimizes Loss 2.

Appendix D contains more details on the experimental setup, factor graph models, and results.

\begin{figure}[!b]
	\centering
	\includegraphics[width=0.9\columnwidth]{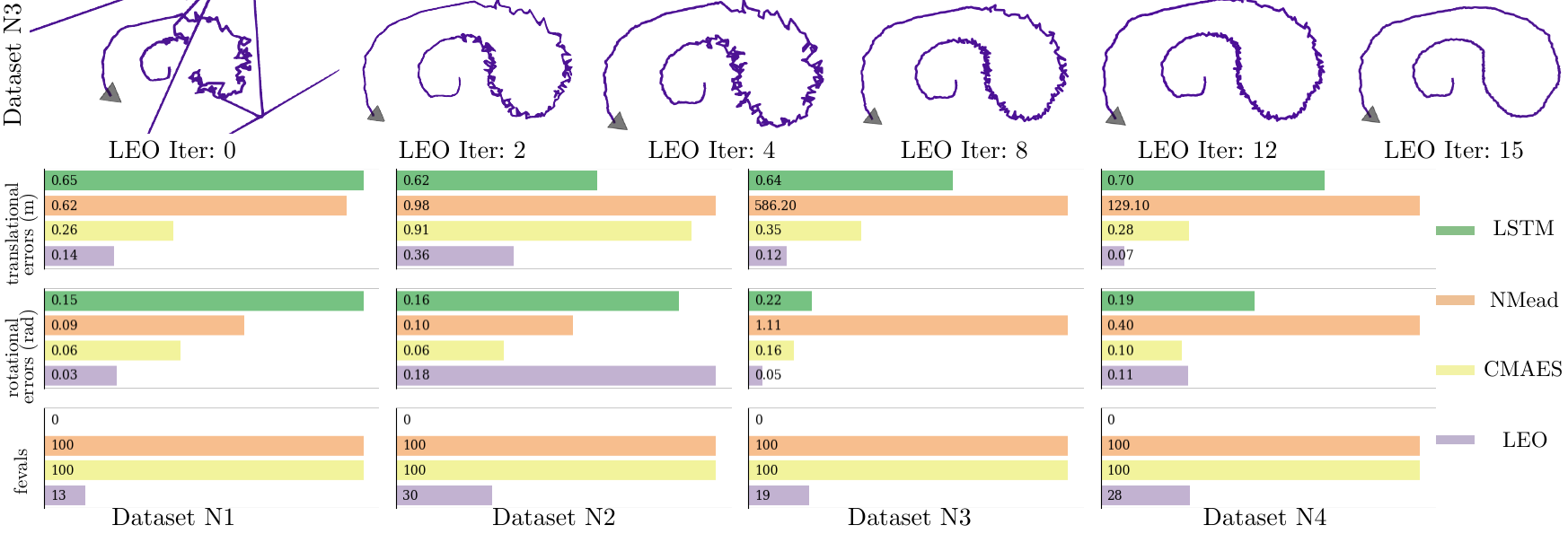}
	\caption{\textbf{Synthetic Navigation Datasets}: Comparison of LEO against baselines. Each dataset N1-N4 consists of $50$ varying trajectories with a 30/20 train/test split. Translational and rotational trajectory tracking errors represent average RMSE over the test set. fevals is number of graph optimizer calls and is computed per data point during training. Runtime $\tiny\sim$$10$ms per optimizer call for a 900-dim state space (3-dof * 300 steps).}
	\figGap
\label{fig:experiment:nav2Dbarplot}
\end{figure}

\vspace{-3mm}
\subsection{End-to-end Tracking Errors}
\vspace{-2mm}
Fig.~\ref{fig:experiment:nav2Dbarplot} shows final tracking results for navigation datasets. We initialize LEO with random $\theta$ such that the graph optimizer returns trajectories far from the ground truth. At the final iteration, LEO converges to a $\theta$ that matches the optimized trajectories to ground truth. We summarize quantitative performance against baselines minimizing final tracking loss (Loss 1). LEO consistently outperforms baselines on final rmse tracking errors. The difference is more pronounced for datasets N3, N4 where the varying covariance model has a higher dimensional parameter space.

\begin{figure}[!t]
	\centering
	\includegraphics[width=\columnwidth]{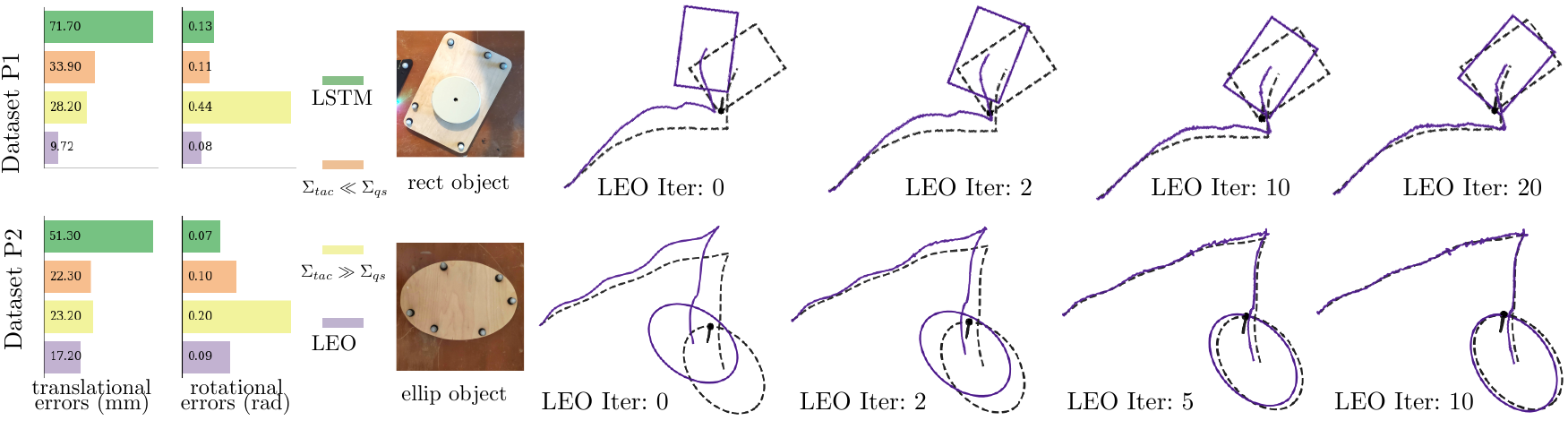}
	\caption{\textbf{Real-world Planar Pushing Datasets}: Each dataset P1, P2 consists of $15$ varying trajectories with a 10/5 train/test split for rect, ellipse object respectively. Trajectory tracking errors represent average RMSE over the test set. Runtime $\tiny\sim$ $50$ms per optimizer call for a 900-dim state space (3-dof * 300 steps).}
	\figGap
\label{fig:experiment:push2Dbarplot}
\end{figure}

Fig.~\ref{fig:experiment:push2Dbarplot} shows final tracking results for real-world planar pushing datasets. Similar to navigation, we show LEO converging to ground truth trajectory for two different objects, a rectangle and ellipse, with differing local contact patches. We report quantitative performance on datasets P1, P2. LEO outperforms different hand-tuned choices of covariances as used in prior work \cite{sodhi2021tactile}. 

\vspace{-3mm}
\subsection{Sample Efficiency and Convergence}
\begin{figure}[!t]
	\centering
	\includegraphics[width=\columnwidth]{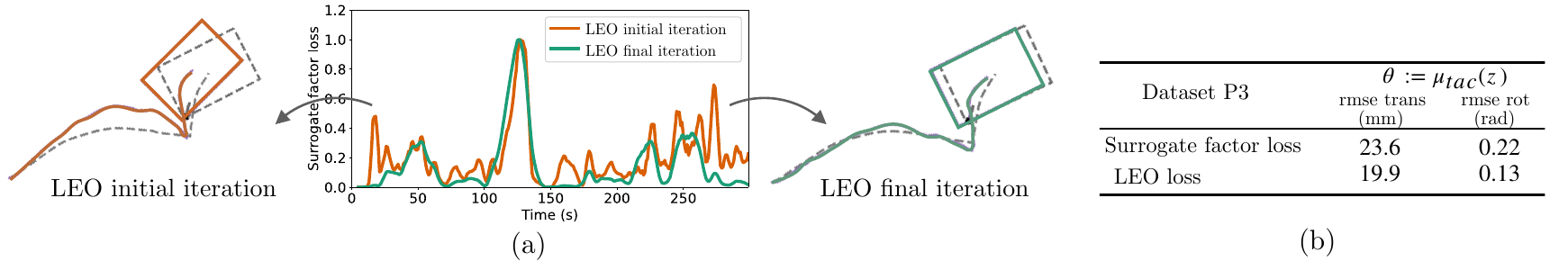}
	\caption{\textbf{(a)} Analysis showing that a surrogate loss, independent of the graph optimizer, may not minimize end-to-end tracking errors. \textbf{(b)} Minimizing only surrogate loss results in a higher final tracking error than minimizing the LEO loss.}
	\figGap
\label{fig:experiment:surrogateLossAnalysis}
\end{figure}

\vspace{-2mm}
Black-box hyper-parameter search is sample inefficient and does not scale well with increasing dimensionality. LEO, on the other hand, is efficient as it uses gradients. This can be seen in Fig.~\ref{fig:experiment:nav2Dbarplot}, where LEO consistently has lower graph optimizer calls (fevals). Moreover, LEO can also be used to learn models with larger parameter spaces, e.g.\ Fig.~\ref{fig:experiment:surrogateLossAnalysis} learns parameters for a tactile factor neural network mapping keypoint image features to transformations (Fig.~\ref{fig:experimentalSetupPlanarPushing}).

\begin{figure}[!t]
	\centering
	\includegraphics[width=0.9\columnwidth]{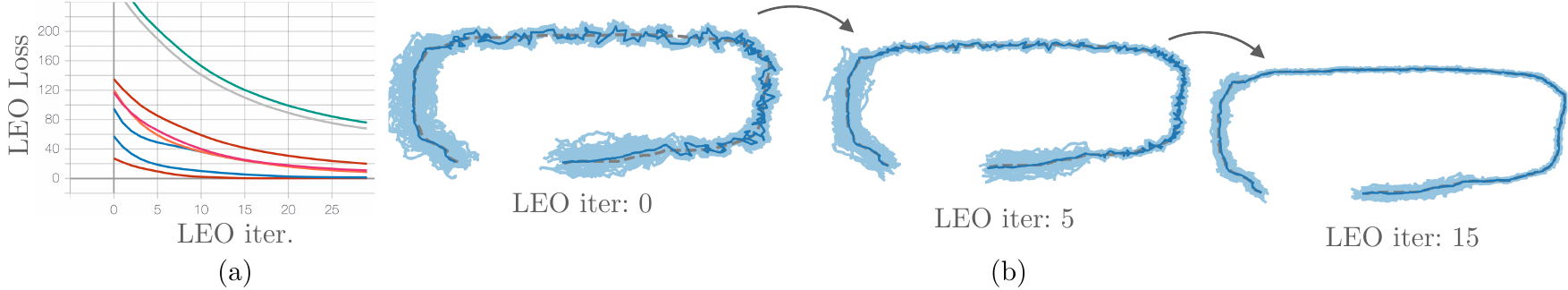}
	\caption{\textbf{(a)} LEO convergence for varying initial $\theta$ on dataset N1. \textbf{(b)} Evolution of graph optimizer trajectory samples over LEO iterations.}
	\figGap
\label{fig:experiment:leoConvergence}
\end{figure}

Finally, Fig.~\ref{fig:experiment:leoConvergence} shows LEO consistently converging for varying $\theta$ initializations on navigation dataset N1. We further show how the LEO samples evolve over iterations. Initially, when the cost is incorrect, the samples are more spread out. Over iterations, as the cost converges, the samples concentrate about the ground truth trajectory.

\vspace{-3mm}
\subsection{Surrogate Loss Analysis}
\vspace{-2mm}
We further analyze why a surrogate loss (Loss 2), independent of the graph optimizer, may not minimize end-to-end tracking errors. We train a network on surrogate supervised learning loss to directly map a pair of tactile image features to ground truth poses similar to prior work~\cite{sodhi2021tactile}. Fig. \ref{fig:experiment:surrogateLossAnalysis}(b) shows the per time step surrogate loss on two trajectories --- an initial trajectory that deviates from ground truth, and a final converged trajectory. While the tracking errors of final is much less than initial, the surrogate loss does not strictly improve. This is further reflected in the table in Fig. \ref{fig:experiment:surrogateLossAnalysis}(c) where the surrogate loss baseline has higher final tracking errors over LEO for dataset P1.

\vspace{-3mm}
\section{Conclusion} 
\label{sec:discussion}
\vspace{-3mm}

We presented an approach, LEO, for learning observation models end-to-end with graph optimizers that may be non-differentiable. We compared LEO against baselines for different observation model types on two distinct tasks of navigation and real-world planar pushing. Overall, LEO is able to learn observation models with lower errors and fewer samples. While we investigated LEO with non-differentiable optimizers in this paper, as future work, we would like to apply LEO to differentiable optimizers and compare against alternate solutions such as unrolling (see Appendix A for a more detailed discussion). We would also like to scale to more complex observation models with applications to 3D object tracking using tactile sensors.

\balance


\footnotesize
\section*{Acknowledgements}
\label{sec:acknowledgement}
This work was supported through the FRAIM program. We thank Frank Dellaert for insightful feedback and suggestions on the paper.
\bibliography{references}

\newpage

\normalsize
\appendix


\section*{Appendix}

\section{Applying LEO to Differentiable Optimizers}
\label{sec:appendix:differentiable}

We discuss here how LEO can be applied to differentiable optimization libraries. It requires two simple changes, (a) disable gradients from the inner loop optimizer block, and (b) replace the direct final tracking loss (Loss 1) with an energy-based loss (Loss 3).  To illustrate how LEO performs compared to alternate solutions such as unrolling, we choose the uni-dimensional regression task from \cite{amos2020differentiable}. This enables us to visualize the learned 2D energy surfaces.

\textbf{Setup} In the regression task, the goal is to learn a network that maps points $(x,y)$ to an energy value, given samples from a ground truth function. The ground truth data is generated from the function $g(x)=x\sin(x)$ for $x\in[0, 2\pi]$. We model $P(y|x)\propto \exp\{-E(\theta, y; x)\}$ where $E(\theta, y; x)=||f(\theta, y; x)||^{2}_2$. $E(.)$ is the energy function expressed as a sum-of-squares of $f(\theta, y; x)$, similar to Section 3.1, to apply Gauss-Newton solvers. $f(\theta, y; x)$ is modeled as a neural network with $\theta$ as the learnable network parameters.

LEO minimizes an energy-based loss (Loss 3), i.e.\ $\mathcal{L}(E(\theta, \cdot); y_{gt}^{i})$, resulting in a similar update rule of the form,
\begin{equation}
	\begin{split}
		\label{eq:appx:eqC.1}
		\scriptsize
		\theta^{+} \leftarrow \theta - \frac{1}{|\mathcal{D}|}\sum^{|\mathcal{D}|}_{i=1}\left\{{\nabla_\theta E(\theta; y^{i}_{gt})-\frac{1}{S}\sum_{\hat{y}} \nabla_{\theta} E(\theta; \hat{y}^{i})}\right\}
	\end{split}
\end{equation}
We solve for this update using the Adam optimizer in Pytorch. We use a batch Gauss-Newton solver as our underlying inner loop optimizer (LEO GN). 

The unrolling baselines minimize a final tracking loss (Loss 1), i.e.\ $\mathcal{L}(\theta; y_{gt})=\nicefrac{1}{|\mathcal{D}|}\sum_{i}||\hat{y}^{i}\ominus y^{i}_{gt}||^2_2$, where $\hat{y}:=\underset{y}{\operatorname{argmin}}\ ||f(\theta, y; x)||^{2}_2$. We use both unrolled gradient descent (Unrolled GD) and unrolled Gauss-Newton (Unrolled GN)~\citep{yi2021differentiable}. 

For all methods, we use two initialization schemes for the underlying optimizer. The first scheme initializes from $y=0$ for all $x$, as followed in~\cite{amos2020differentiable}. The second scheme initializes from the ground truth $g(x)=x\sin(x)$ as suggested in~\cite{yi2021differentiable}.

\begin{figure}[!b]
	\centering
	\includegraphics[width=\textwidth, trim = 0cm 0cm 0cm 0cm]{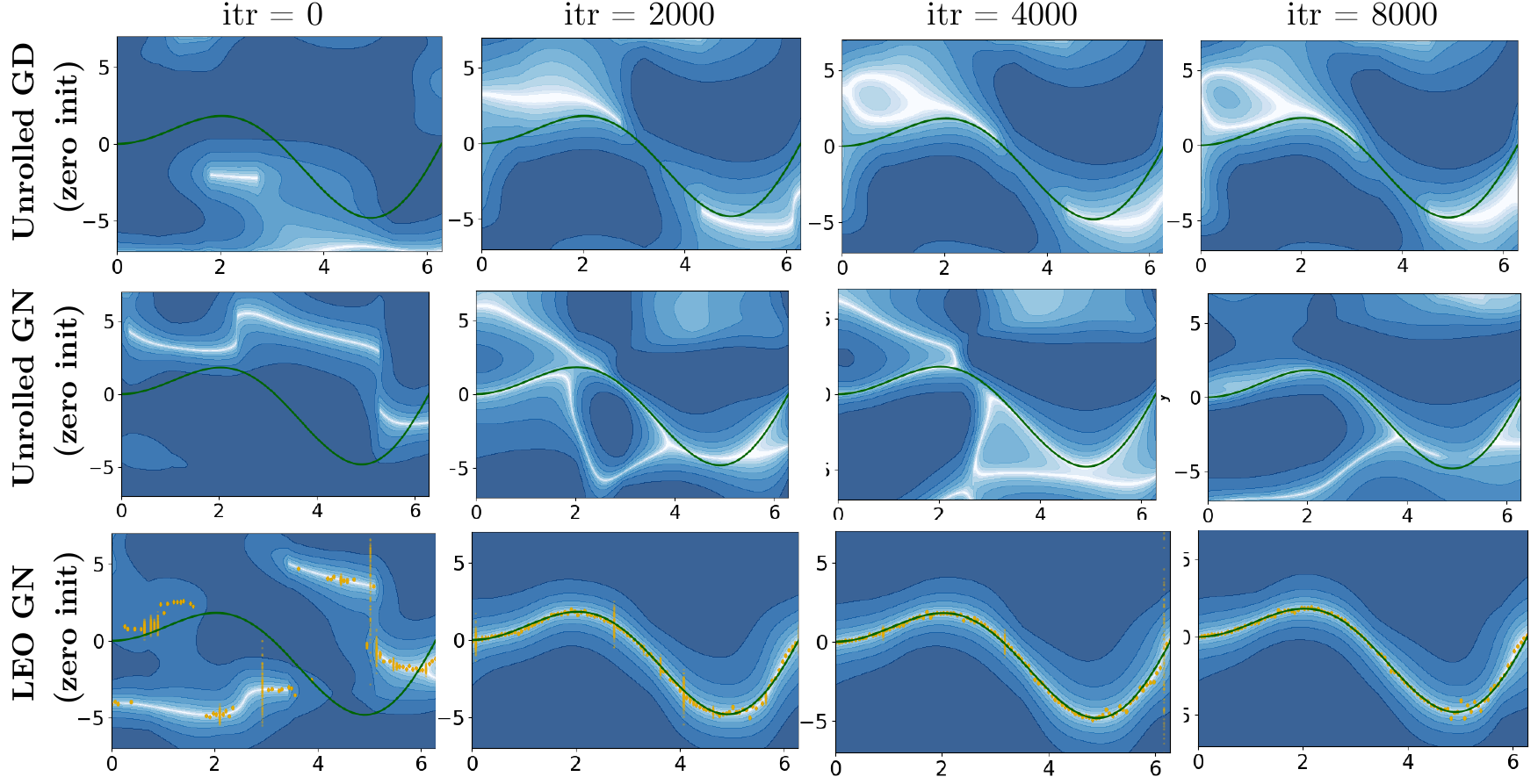}
	\caption{Evolution of learned 2D energy surfaces using \textbf{initialization scheme 1}. Contour surfaces show the normalized energy surfaces. Ground truth function $x\sin(x)$ in green, LEO samples in orange. Lighter colors correspond to lower energy.}
	\label{fig:regressionEnergySurfaceZeroInit}
\end{figure}

\begin{figure}[!t]
	\centering
	\includegraphics[width=\textwidth, trim = 0cm 0cm 0cm 0cm]{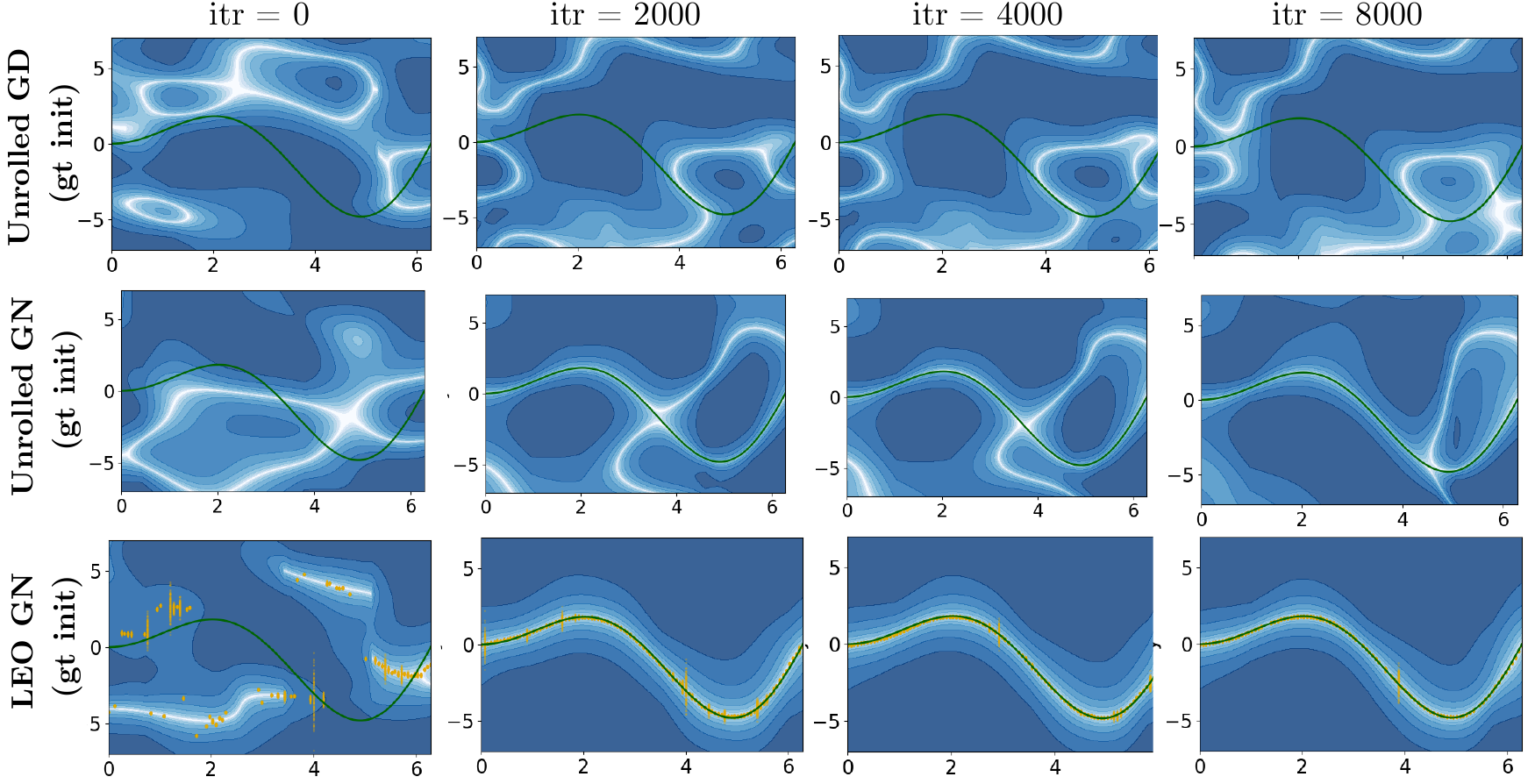}
	\caption{Evolution of learned 2D energy surfaces using \textbf{initialization scheme 2}. Contour surfaces show the normalized energy surfaces. Ground truth function $x\sin(x)$ in green, LEO samples in orange. Lighter colors correspond to lower energy.}
	\label{fig:regressionEnergySurfaceGTInit}
\end{figure}

\textbf{Results}
Fig.~\ref{fig:regressionEnergySurfaceZeroInit} shows the learned energy surfaces for all 3 methods using initialization scheme 1, i.e., initializing the optimizer from zero. At iteration $0$, all 3 approaches are initialized with a random energy function. While all 3 approaches end up with a low final loss, only LEO converges to an energy function with a single basin about the ground truth samples. The unrolled baselines converge to energy functions with multiple minima, not all of them on the ground truth samples. This is because the energy function need only be good enough for the optimizer, when initialized from $y = 0$ to reach the ground truth for a fixed number of unrolling steps. It does not guarantee a similar convergence when executed with different number of unrolling steps or initialization from different $y$ values.

Fig.~\ref{fig:regressionEnergySurfaceGTInit} shows the learned energy surfaces for initialization scheme 2, i.e., initializing the optimizer from ground truth. Similar to the previous figure, all 3 methods initialize from random energy functions and converge to a low final loss. But only LEO converges to a similar energy function as before, i.e., one with a single basin around the ground truth. The unrolling baselines converge to a different energy function than in scheme 1. In this case, the energy function need only be good enough for the optimizer, when initialized from ground-truth, to stay on the ground truth.

For both schemes, LEO alone converges to the same energy function, as it is less sensitive to the underlying optimization process.  
In future work, we aim to apply LEO on differentiable optimizers for robot estimation tasks. This would require a factor graph optimization library similar to GTSAM but implemented in a differentiable manner. There are recent efforts on this front~\citep{yi2021differentiable,swiftfusion2021,caesarjl2021}, but there isn't a single, stable open-source library yet.  We hope our analysis on this regression task would encourage others in the community to try out LEO in their respective differentiable optimizer libraries.



\section{Comparison to a MCMC sampler}
\label{sec:appendix:mcmc}

\begin{wrapfigure}{l}{0.4\textwidth}
	\centering
	\includegraphics[width=0.4\textwidth, trim = 0cm 0cm 0cm 0cm]{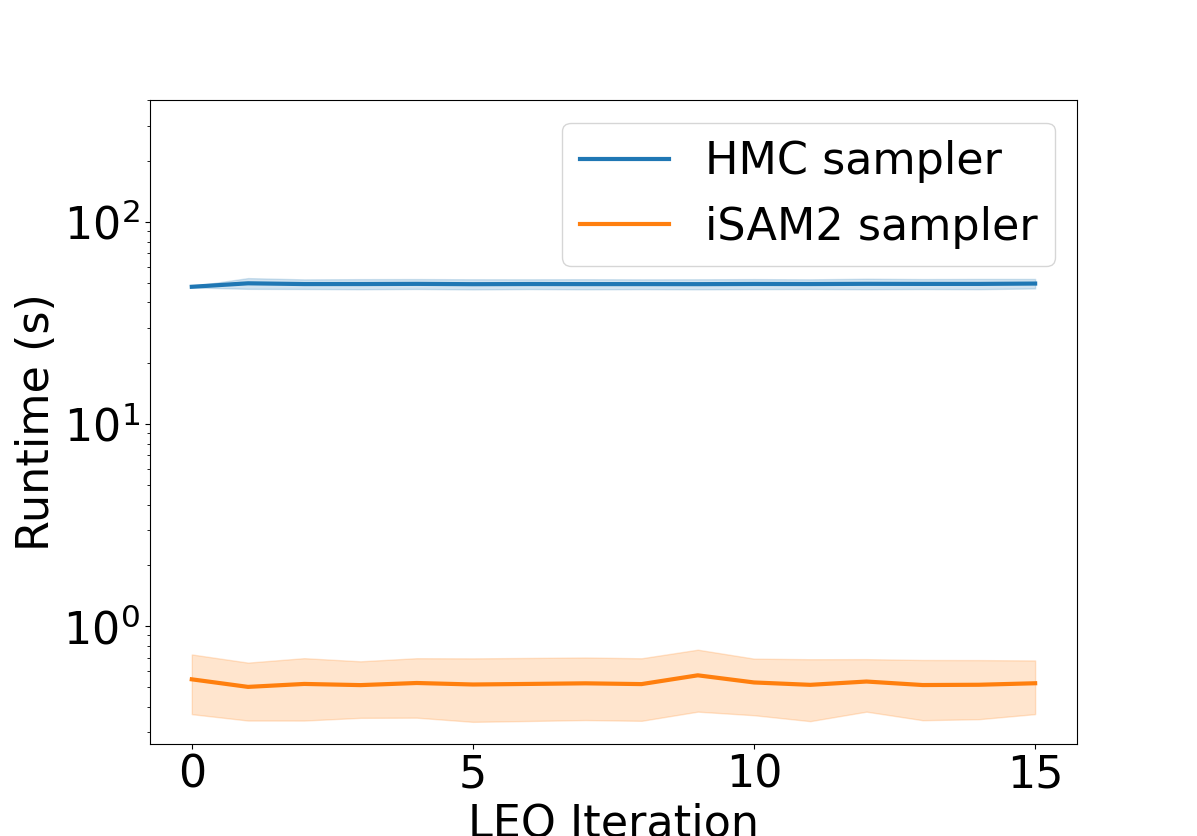}
	\caption{Runtime comparison against an HMC sampler}
	\label{fig:samplerRuntime}
\end{wrapfigure}

We compare run times against an HMC sampler implemented in the hamiltorch library \cite{cobb2020scaling}. The sampler simulates a set of Hamiltonian differential equations which upon simulating generates a sequence of samples.

We run this on the 2D navigation dataset which have a 900-dimensional state space (3-DOF * 300 steps). The HMC sampler is sampling from the true posterior distribution while the incremental Gauss-Newton, iSAM2 \cite{kaess2012isam2}, sampler that we use samples from a Gaussian approximation that it maintains online. However, we found similar convergence at train time using both samplers which suggests that the Gaussian approximation was reasonable for our applications.

In terms of run time, we found the HMC sampler run time to be two orders of magnitude greater than the iSAM2 sampler as shown in Fig. \ref{fig:samplerRuntime}. This is expected since the HMC sampler evaluates the cost as well the gradients of the cost every time a sample is generated which requires looping through all the factors and can be expensive. In contrast, the iSAM2 sampler allows for directly sampling from the Gaussian that is significantly faster. When used in an online setting, as new factors get added to the graph, the HMC sampler would have to recompute the posterior distribution. On the other hand, the iSAM2 sampler efficiently and incrementally updates the Gaussian approximation by leveraging sparsity.

\section{Principle of Maximum Entropy}
\label{sec:appendix:maxent}

In Section 4.1, we defined the energy-based loss as the normalized negative log-likelihood loss. The motivation for such a loss comes from probabilistic modeling, notably the principle of maximum entropy moment matching.

\begin{wrapfigure}{!r}{0.5\textwidth}
	\centering
	\includegraphics[width=0.5\textwidth, trim = 0cm 0cm 0cm 0cm]{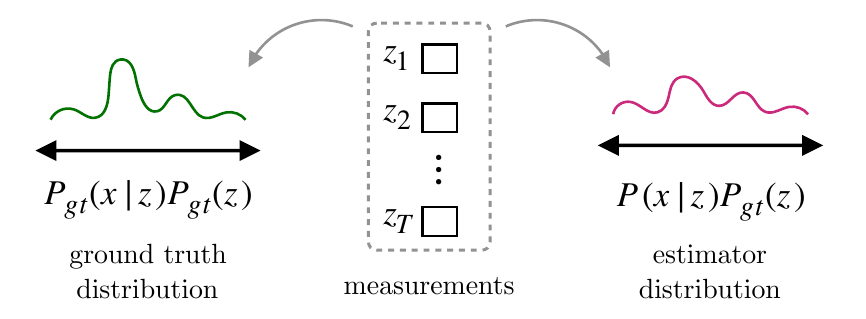}
	\caption{Ground truth distribution is unknown and only observable via measurements $z_{1:t}$. The goal is to estimate a distribution $P(x|z)$ that matches observed moments.}
	\label{fig:distributionMatching}
\end{wrapfigure}

Assume a ground truth distribution $P_{gt}(x,z)$ that generates x and z. We would like to estimate a distribution $P_{\theta}(x|z)$ that matches this.  The true ground truth distribution is unknown, but we can sample from it. This allows us to estimate $P_{\theta}(x|z)$ by matching empirical moments computed from these samples (Fig. \ref{fig:distributionMatching}).

However, there is a continuous space of  probability distributions that match these empirical moments. Which of these distributions do we pick? The maximum entropy (MaxEnt) principle suggests we pick the distribution which is least committal, i.e.\ the one with maximum entropy \cite{jaynes1957information}. We can write out this objective of maximizing entropy of the distribution $P_{\theta}(x|z)$ subject to matching moment constraints against the unknown ground truth distribution as,
\begin{equation}
	\begin{split}
	\label{eq:appx:eqA.1}
	& \operatorname{max}\ -\sum\limits_{x}\ P_{\theta}(x|z)\log P_{\theta}(x|z) \\ 
	s.t. \underset{{{\substack{z\sim P_{gt}}(z)}}}{\operatorname{E}} & \underset{{{\substack{x\sim P_{\theta}(x|z)}}}}{\operatorname{E}} \Phi(x,z) = \underset{{{\substack{(x,z)\sim P_{gt}}(x,z)}}}{\operatorname{E}} \Phi(x,z)
	\end{split}
\end{equation}
where, $\Phi(\cdot)$ are the moment functions. Applying KKT conditions for Eq. \ref{eq:appx:eqA.1} and expanding the dual results in an unconstrained objective of the form \cite{jaynes1957information,ziebart2010modeling} (more details in Eq. \ref{eq:end2end_learnopt:appx:moments:eq1}),
\begin{equation}
	\begin{split}
	\label{eq:appx:eqA.2}
	& \underset{{\theta}}{\operatorname{max}}\ \underset{{\substack{x,z\sim \\ P_{gt}(x,z)}}}{\operatorname{E}}\ \log P_{\theta}(x|z) \\
	& P_{\theta}(x|z) \propto \exp(-\theta^T\Phi(x,z))
\end{split}
\end{equation}

where, $\theta$ are the Lagrange multipliers. The term $\theta^{T}\Phi(x,z)=E(\theta, x;z)$ can be interpreted as a linear ``energy function'' such that $P_{\theta}(x)\propto\exp(-E(\theta, x;z))$ puts a high energy on low probability solutions. This linear class of cost functions can be lifted to a richer class of parametric nonlinear functions $E(\theta, x;z)$. Note that this may induce a duality gap in optimizing Eq. \ref{eq:appx:eqA.1}, but is commonly used in practice when parameterizing energy functions as neural networks \cite{lecun2006tutorial, levine2011nonlinear, finn2016connection}. The maximum entropy solution can then be written out as,
\begin{equation}
	\begin{split}
	\label{eq:appx:eqA.3}
	& \underset{{\theta}}{\operatorname{max}}\ \underset{{\substack{x,z\sim \\ P_{gt}(x,z)}}}{\operatorname{E}}\ \log P_{\theta}(x|z) \\
	& P_{\theta}(x|z) \propto \exp(-E({\theta},x;z))
	\end{split}
\end{equation}
Eq. \ref{eq:appx:eqA.3} asks us to maximize the likelihood of ground truth trajectories where the likelihood takes the form of a Boltzmann distribution, where lower energy implies higher probability mass. In other words, we want to find energy function $E({\theta},x;z)$ that places maximal probability mass on the ground truth trajectory. Writing out this objective over a training dataset $(x_{gt},z)\in\mathcal{D}$ results in the NLL loss that we saw earlier in Section 4.1,
\begin{equation}
	\begin{split}
	\label{eq:appx:eqA.4}
	\mathcal{L}(\theta) & =  \frac{1}{|\mathcal{D}|}\sum_{(x_{gt}^{i},z^{i})\in \mathcal{D}} -\log P_{\theta}(x_{gt}^{i}|z^{i}) \\
\end{split}
\end{equation}

Let's expand out the steps in Section 4.1 in more detail. Substituting the expression for Boltzmann distribution $P_{\theta}(x|z)$ results in,
\begin{equation}
	\begin{split}
	\label{eq:appx:eqA.5}
	\mathcal{L}(\theta) & =  \frac{1}{|\mathcal{D}|}\sum_{(x_{gt}^{i},z^{i})\in \mathcal{D}}  -\log (\exp(-E(\theta; x_{gt}^{i}, z^{i}))) + \log Z(\theta; z^{i}) \\
	\mathcal{L}(\theta) & =  \frac{1}{|\mathcal{D}|}\sum_{(x_{gt}^{i},z^{i})\in \mathcal{D}} E(\theta; x_{gt}^{i}, z^{i}) + \log \int_x \exp(-E(\theta; x, z^{i}))dx
	\end{split}
\end{equation}

Take the gradient of this loss, 
\begin{equation}
	\begin{split}
	\label{eq:appx:eqA.6}
	\nabla_{\theta} \mathcal{L}(\theta) & = \frac{1}{|\mathcal{D}|}\sum_{(x_{gt}^{i},z^{i})\in \mathcal{D}} \left[ \nabla_{\theta} E(\theta; x_{gt}^{i}, z^{i})) + \frac{1}{Z(\theta; z^{i})}\int_x\nabla_{\theta}E(\theta;x,z^{i})\exp(-E(\theta;x,z^{i}))\right] \\
	\end{split}
\end{equation}

Substituting in $P_{\theta}(x|z^{i})$ from Section 3.1 we have the resulting gradient expression that we finally obtained in Section 4.1,
\begin{equation}
	\begin{split}
	\label{eq:approach:eq4.1.9}
	\nabla_{\theta} \mathcal{L}(\theta) = \frac{1}{|\mathcal{D}|}\sum_{(x_{gt}^{i},z^{i})\in \mathcal{D}} \left[ \underbrace{\nabla_{\theta} E(\theta; x_{gt}^{i}, z^{i}))}_\text{ground truth samples} - \underset{\substack{x\sim \\ P_{\theta}(x|z)}}{\operatorname{\mathbb{E}}}\underbrace{\nabla_{\theta} E(\theta; x, z^{i}))}_\text{learned distribution samples}\right]
	\end{split}
\end{equation}

\textbf{Negative log likelihood approximation}

Since directly sampling from the true posterior $P_{\theta}(x|z)=\frac{1}{\mathcal{Z}(\theta)}e^{-E(\theta,x;z)}$ is intractable, in LEO, we instead sample from a Gaussian approximation of this distribution, $\hat{P}_{\theta}(x|z)=\mathcal{N}(\mu, \Sigma)$ where $\mu$, $\Sigma$ are obtained as shown in Section \ref{sec:approach:gaussnewton}. Hence, we effectively minimize an approximation of the true NLL loss,
\begin{equation}
	\begin{split}
	\label{eq:appx:eqA.2.2}
	\hat{\mathcal{L}}(\theta) & =  \frac{1}{|\mathcal{D}|}\sum_{(x_{gt}^{i},z^{i})\in \mathcal{D}} -\log \hat{P}_{\theta}(x_{gt}^{i}|z^{i})
\end{split}
\end{equation}

The difference between the approximated loss and the true loss is,
\begin{equation}
	\begin{split}
	\label{eq:appx:eqA.2.3}
	\hat{\mathcal{L}}(\theta) - \mathcal{L}(\theta) & =\frac{1}{|\mathcal{D}|}\sum_{(x_{gt}^{i},z^{i})\in \mathcal{D}} -\log \hat{P}_{\theta}(x_{gt}^{i}|z^{i}) - \frac{1}{|\mathcal{D}|}\sum_{(x_{gt}^{i},z^{i})\in \mathcal{D}} -\log {P}_{\theta}(x_{gt}^{i}|z^{i}) \\
	& =\frac{1}{|\mathcal{D}|}\sum_{(x_{gt}^{i},z^{i})\in \mathcal{D}} \frac{P_{\theta}(x_{gt}^{i}|z^{i})}{\hat{P}_{\theta}(x_{gt}^{i}|z^{i})} \\
	& = \underset{\substack{(x_{gt},z)\sim \mathcal{D}}}{\operatorname{\mathbb{E}}}\ \ \frac{P_{\theta}(x_{gt}|z)}{\hat{P}_{\theta}(x_{gt}|z)} \ \leq \max_{\substack{(x_{gt},z)\in \mathcal{D}}} \ \log \frac{P_{\theta}(x_{gt}|z)}{\hat{P}_{\theta}(x_{gt}|z)} 
\end{split}
\end{equation}
The difference in losses is bounded by the maximum log density ratio between the true and approximated distributions. The ratio is bounded as long as the denominator is not zero, i.e. the approximation has full support over the dataset.

\textbf{Derivation from moment matching}

When going from Eq. \ref{eq:appx:eqA.1} to \ref{eq:appx:eqA.2}, we rewrote the constrained maximum entropy objective as an unconstrained objective. Let's derive this for our setting \cite{jaynes1957information}. For simplicity of notations, we'll drop the conditional $z$ in the probability expressions.

We would like to maximize (or minimize negative) entropy subject to matching moments, i.e.,
\begin{equation}
\begin{split}
    \label{eq:end2end_learnopt:appx:moments:eq1}
    & \min_x\sum_{x}P_{\theta}(x)\log P_{\theta}(x) \\
    \text{s.t., }\ \ \ \ \ & \sum_{x} P_{\theta}(x) \Phi(x) = \sum_{x} P_{gt}(x) \Phi(x) = c \hspace{10pt} \textit{(moment constraint)}\\
    & \sum_{x} P_{\theta}(x) = 1 \hspace{90pt} \textit{(probability sum constraint)} \\
\end{split}
\end{equation}

Writing out the Lagrangian for this constrained problem,
\begin{equation}
    \begin{split}
        \label{eq:end2end_learnopt:appx:moments:eq2}
        \mathcal{L}(\theta, \lambda, \mu)= \sum_{x}P_{\theta}(x)\log P_{\theta}(x) + \lambda \left(\sum_{x} P_{\theta}(x)\Phi(x) - c\right) + \mu \left(\sum_{x} P_{\theta}(x) - 1\right)
    \end{split}
\end{equation}
where, $\lambda$, $\mu$ are the Lagrange multipliers. Taking gradient of Lagrangian and setting to $0$,
\begin{equation}
    \begin{split}
        \label{eq:end2end_learnopt:appx:moments:eq3}
        & \hspace{50pt} \nabla_{\theta} \mathcal{L}(\theta, \lambda, \mu) = 0 \\ 
        \Rightarrow & \sum_{x}\cancel{{P_{\theta}(x)}}\frac{1}{\cancel{P_{\theta}(x)}}\nabla P_{\theta}(x) + \sum_{x} \log P_{\theta}(x) \nabla_{\theta}P_{\theta}(x) + \\ 
        & \hspace{50pt} \lambda \sum_{x} \nabla_{\theta} P_{\theta}(x)\Phi(x) +
        \mu \sum_{x} \nabla_{\theta} P_{\theta}(x) = 0 \\
        \Rightarrow & \sum_x \nabla P_{\theta}(x)(1+\log P_{\theta}(x)+\lambda\Phi(x)+\mu) = 0
    \end{split}
\end{equation}

Eq. \ref{eq:end2end_learnopt:appx:moments:eq3} goes to $0$ for a probability distribution $P_{\theta}(x)$ of the form,
\begin{equation}
    \begin{split}
        \label{eq:end2end_learnopt:appx:moments:eq4}
        P_{\theta}(x)=\exp^{-(1+\mu+\lambda\Phi(x))}=\frac{1}{{Z}(\theta,\lambda)}\exp(-\lambda\Phi(x))
    \end{split}
\end{equation}
where, ${Z}(\theta,\lambda)=\sum_x\exp(-\lambda\Phi(x))$ such that $\sum_xP_{\theta}(x)=1$. We can subsume $1+\mu$ since that's a constant.

Writing out the dual form $\max_{\lambda}\mathcal{L}(\theta,\lambda,\mu)$ of the Lagrangian with $P_{\theta}(x)$ expression substituted from Eq. \ref{eq:end2end_learnopt:appx:moments:eq4},
\begin{equation}
    \begin{split}
        \label{eq:end2end_learnopt:appx:moments:eq5}
        \hspace{30pt} \max_{\lambda}\ \ & \sum_x \frac{1}{Z(\theta, \lambda)}\exp\left(-\lambda\Phi(x)\right)\log\left[\frac{1}{Z(\theta, \lambda)}\exp(-\lambda\Phi(x))\right] + \\ 
        & \lambda \left[\sum_x \frac{1}{Z(\theta, \lambda)}\exp(-\lambda\Phi(x))\Phi(x)-c\right] + \mu \cancel{\left[\sum_x \frac{1}{Z(\theta, \lambda)}\exp(-\lambda\Phi(x))\Phi(x)-1\right]} \\
        \Rightarrow \hspace{30pt} \max_{\lambda}\ \ 
        & \cancel{\sum_x \frac{1}{Z(\theta, \lambda)}\exp\left(-\lambda\Phi(x)\right)\left[-\lambda\Phi(x)\right]} + \sum_x \frac{1}{Z(\theta, \lambda)}\exp\left(-\lambda\Phi(x)\right)\left[-\log Z(\theta, \lambda)\right] + \\ 
        & \cancel{\lambda \sum_x \frac{1}{Z(\theta, \lambda)}\exp\left(-\lambda\Phi(x)\right)\Phi(x)} - \lambda c \\
        \Rightarrow \hspace{30pt} \max_{\lambda}\ \ & -\log Z(\theta, \lambda) \left(\sum_x \frac{1}{Z(\theta, \lambda)}\exp\left(-\lambda \Phi(x)\right)\right) - \lambda c \\
        \Rightarrow \hspace{30pt} \max_{\lambda}\ \ & -\log Z(\theta, \lambda) - \lambda c
    \end{split}
\end{equation}

Introducing true distribution $P_{gt}(x)$ and rearranging terms results in the unconstrained objective that we saw in Eq. \ref{eq:appx:eqA.2},
\begin{equation}
    \begin{split}
        \label{eq:end2end_learnopt:appx:moments:eq6}
        & \max_{\lambda}\ \ -\log Z(\theta, \lambda) - \lambda c \\
        \Rightarrow \hspace{30pt} & \max_{\lambda}\ \ \sum_x P_{gt}(x) \left[-\log Z(\theta, \lambda)\right] - \lambda \sum_x P_{gt}(x)\Phi(x) \\
        \Rightarrow \hspace{30pt} & \max_{\lambda}\ \ \sum_x P_{gt}(x) \left[-\log Z(\theta, \lambda) - \lambda \Phi(x)\right] \\
        \Rightarrow \hspace{30pt} & \max_{\lambda}\ \ \sum_x P_{gt}(x) \log\left[\exp\left(\frac{-\lambda\Phi(x)}{Z(\theta,\lambda)}\right)\right] \\
        \Rightarrow \hspace{30pt} & \max_{\lambda}\ \ \sum_x P_{gt}(x) \log P_{gt}(x)
    \end{split}
\end{equation}



\section{Experimental Evaluation: Additional Details}
\label{sec:appendix:results}

\subsection{Setup and Factor Graph Models}

\begin{wrapfigure}{l}{0.6\textwidth}
	\centering
	\includegraphics[width=0.6\textwidth, trim = 0cm 0cm 0cm 0cm]{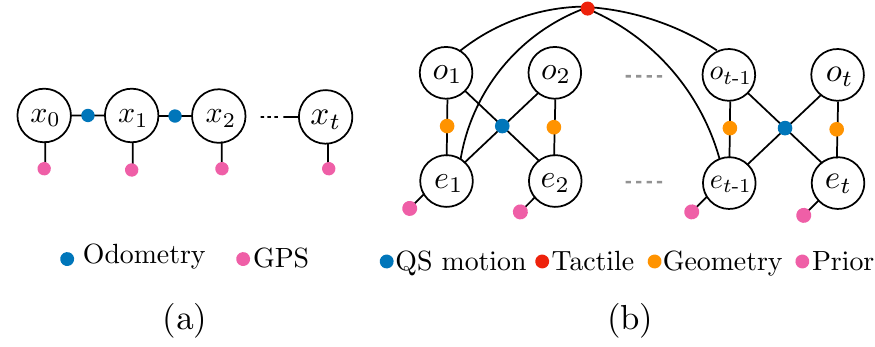}
	\caption{Factor graphs for (a) synthetic navigation (b) real-world planar pushing.}
	\label{fig:datasetsFactorGraphs}
\end{wrapfigure}

Fig. \ref{fig:datasetsFactorGraphs} shows the two factor graphs modeling the (a) synthetic navigation and (b) real-world planar pushing tasks. For both graphs, poses are modeled as variable nodes and measurements as factor nodes. The graph inference objective is to solve for the latent poses (variables) given measurements (factors).


\textbf{Synthetic Navigation}
To replicate a typical robot navigation setup, we generate odometry measurements by adding Gaussian noise to \textit{relative} ground truth state estimates and GPS measurements by adding Gaussian noise to \textit{absolute} ground truth estimates. We generate 4 datasets N1--N4, each with a different covariance setting. N1, N2 has measurements generated from fixed covariances $\Sigma:=\{\Sigma_{odom}, \Sigma_{gps}\}$. N3, N4 has measurements generated from covariances varying between two sets of values, i.e.\ $\Sigma:=\{\Sigma_{odom}(z), \Sigma_{gps}(z)\}$, where $z$ are binary measurements simulating an ambient light detector that determines whether the robot is in an indoor or outdoor environment.

To solve for the sequence of states $x:=\{x_1 \dots x_{T}\}$ from measurements $z$, we provide the following objective for the graph optimizer to minimize,
\begin{equation}
	\begin{split}
		\label{eq:appx:eqB.1}
		\hat{x} = & \underset{x}{\operatorname{argmin}}\sum_{k} \left\{\right.||f_{odom}(x_{k-1}, x_{k})-z^{odom}_{k-1, k}||^2_{\Sigma_{odom}(z)} + ||f_{gps}(x_{k})-z^{gps}_{k}||^2_{\Sigma_{gps}(z)} \left.\right\}
	\end{split}
\end{equation}
The observation model parameters that we learn for this task are the fixed and varying covariances, i.e. $\theta:=\{\Sigma_{odom}, \Sigma_{gps}\}, \{\Sigma_{odom}(z), \Sigma_{gps}(z)\}$.

\textbf{Real-world planar pushing}
For the planar pushing setup, states $x:=\{x_1 \dots x_{T}\}$ in the graph are the planar object and end-effector poses at every time step, with $x_t=[o_t\ e_t]^T$, where $o_t, e_t \in SE(2)$. Factors in the graph incorporate tactile observations $f_{tac}(\cdot)$, quasi-static pushing dynamics $f_{qs}(\cdot)$, geometric constraints $f_{geo}(\cdot)$, and priors on end-effector poses $f_{eff}(\cdot)$.

To solve for the sequence of state $x:=\{x_1 \dots x_{T}\}$ from measurements $z$, we pass in the following objective for the graph optimizer to minimize,
\begin{equation}
	\begin{split}
		\label{eq:appx:eqB.2}
		\scriptsize
		\hat{x} = \underset{x}{\operatorname{argmin}}\sum_{k}^{} & \left\{\right. ||f^{\phi}_{tac}(o_{k-w}, o_{k}, e_{k-w}, e_{k})||^2_{\Sigma_{tac}} + ||f_{qs}(o_{k-1}, o_{k}, e_{k-1}, e_{k})||^2_{\Sigma_{qs}} + \\
		& ||f_{geo}(o_{k}, e_{k})||^2_{\Sigma_{geo}} + ||f_{eff}(e_{k})||^2_{\Sigma_{eff}} \left.\right\}
	\end{split}
\end{equation}
The observation model parameters that we learn for this task are the tactile factor network weights and the tactile and quasi-static factor covariances, i.e.\ $\theta:=\{\phi, \Sigma_{tac}, \Sigma_{qs}\}$. Tactile factor network contains an image to keypoint feature encoder pre-trained using a self-supervised image reconstruction loss and a fully connected network mapping keypoint features to relative transforms. We fine tune weights for the keypoint to relative transform portion of the network and use fixed, pre-trained weights for feature encoder. Tactile factors are added between non-consecutive poses $\{k-w, k\}$ over a window with the window length chosen heuristically, and quasi-static motion factors are added between consecutive poses $\{k-1, k\}$ (Fig. \ref{fig:datasetsFactorGraphs}). More details on the factor graph models along with the network architecture can be found in prior work \cite{sodhi2021tactile}.



\begin{figure}[!t]
	\centering
	\includegraphics[width=\textwidth, trim = 0cm 0cm 0cm 0cm]{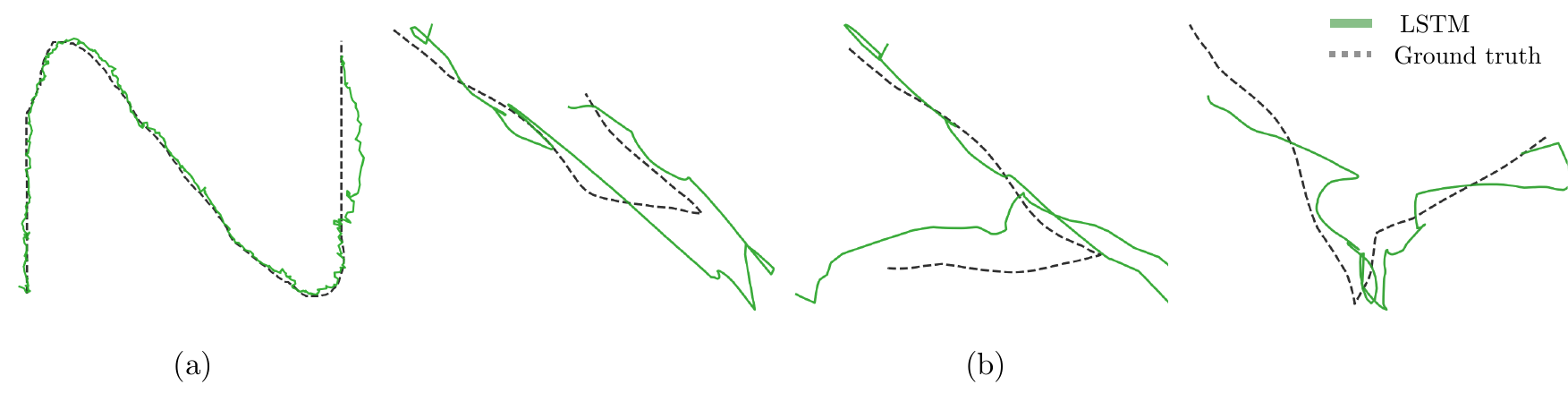}
	\caption{Qualitative LSTM results on \textbf{(a)} a synthetic navigation robot trajectory \textbf{(b)} planar pushing object trajectories.}
	\label{fig:lstmQualPlots}
	\figGap
\end{figure}

\subsection{Baselines}

For the hyper-parameter search baselines we use off-the-shelf solvers like CMA-ES \cite{hansen2019pycma}, and scipy optimizers such as Nelder-Mead. For the learned sequence model baseline, we use an LSTM. 

\textbf{LSTM architecture} For the synthetic navigation task, we directly regresses to absolute poses. At each time step, the absolute pose and odometry measurements are concatenated as inputs, and the network predicts the 2D position and the sin and cos of the rotation angle. We use a 2-layer LSTM with 64 hidden units, followed by a 2 dense layers of 32 hidden units before decoding into the 4D output. The loss is a weighted sum of translation and rotation errors. In the real-world planar pushing datasets, we do not have direct measurements of the object pose. Therefore, we transform all end-effector trajectories to start at the origin, which improves the generalization. Each input consists of the end-effector pose observation and the 2d object contact observation, and the same architecture as the navigation dataset is used. In addition, we found that increasing the sequence length with each epoch improved performance.

Fig. \ref{fig:lstmQualPlots} shows qualitative results for the LSTM on the (a) synthetic navigation and (b) real-world planar pushing datasets. The performance on the planar pushing object trajectories is not as good since, unlike navigation where we had noisy GPS measurements, tactile measurements only give partially observable information about the object state.

\end{document}